\documentclass[a4paper,conference]{IEEEtran}
\usepackage{amsmath}
\usepackage{amssymb}
\usepackage{multirow}
\usepackage{lipsum}
\usepackage{subcaption}
\usepackage{makecell}
\usepackage{booktabs}
\usepackage{times}
\usepackage{epsfig}
\usepackage{graphicx}

\newcommand{\etal}{\textit{et al}. }

\ifCLASSINFOpdf
\else
\fi
\begin{document}
%

\title{EdgeNet: Semantic Scene Completion from a Single RGB-D Image}

\author{Aloisio Dourado, Teofilo Emidio de Campos\\
University of Brasilia\\
Brasilia, Brazil\\
{\tt\small t.decampos@st-annes.oxon.org}
\and
Hansung Kim, Adrian Hilton\\
University of Surrey\\
Surrey, UK\\
{\tt\small (h.kim, a.hilton)@surrey.ac.uk}
}

\author{\IEEEauthorblockN{Aloisio Dourado, Teofilo Emidio de Campos}
\IEEEauthorblockA{University of Brasilia\\
Brasilia, Brazil\\
{aloisio.dourado.bh@gmail.com, t.decampos@st-annes.oxon.org}}
\and
\IEEEauthorblockN{Hansung Kim, Adrian Hilton}
\IEEEauthorblockA{University of Surrey\\
Surrey, UK\\
{(h.kim, a.hilton)@surrey.ac.uk}}
}


%


\maketitle

\begin{abstract}
Semantic scene completion is the task of predicting a complete 3D representation of volumetric occupancy with corresponding semantic labels for a scene from a single point of view. In this paper, we present EdgeNet, a new end-to-end neural network architecture that fuses information from depth and RGB, explicitly representing RGB edges 
in 3D space. Previous works on this task used either depth-only or depth with colour by projecting 2D semantic labels generated by a 2D segmentation network into the 3D volume, requiring a two step training process. 
Our EdgeNet representation encodes colour information in 3D space using edge detection and flipped truncated signed distance, which improves semantic completion scores  especially in hard to detect classes. We achieved state-of-the-art scores on both synthetic and real datasets with a simpler and a more computationally efficient training pipeline than competing approaches. 
\end{abstract}


%
\IEEEpeerreviewmaketitle

\section{Introduction}
The ability of reasoning about scenes in 3D is a natural task
for humans, but remains a challenging problem in Computer Vision \cite{marr_vision:_1982}. 
Knowing the complete 3D geometry of a scene and the semantic labels of each 3D voxel has 
many practical applications,
like  robotics and autonomous navigation in indoor environments,  surveillance,  
assistive  computing and augmented reality. 

\begin{figure}[t]
\begin{center}
   \includegraphics[trim=0 85 105 0, clip, width=0.9\linewidth]{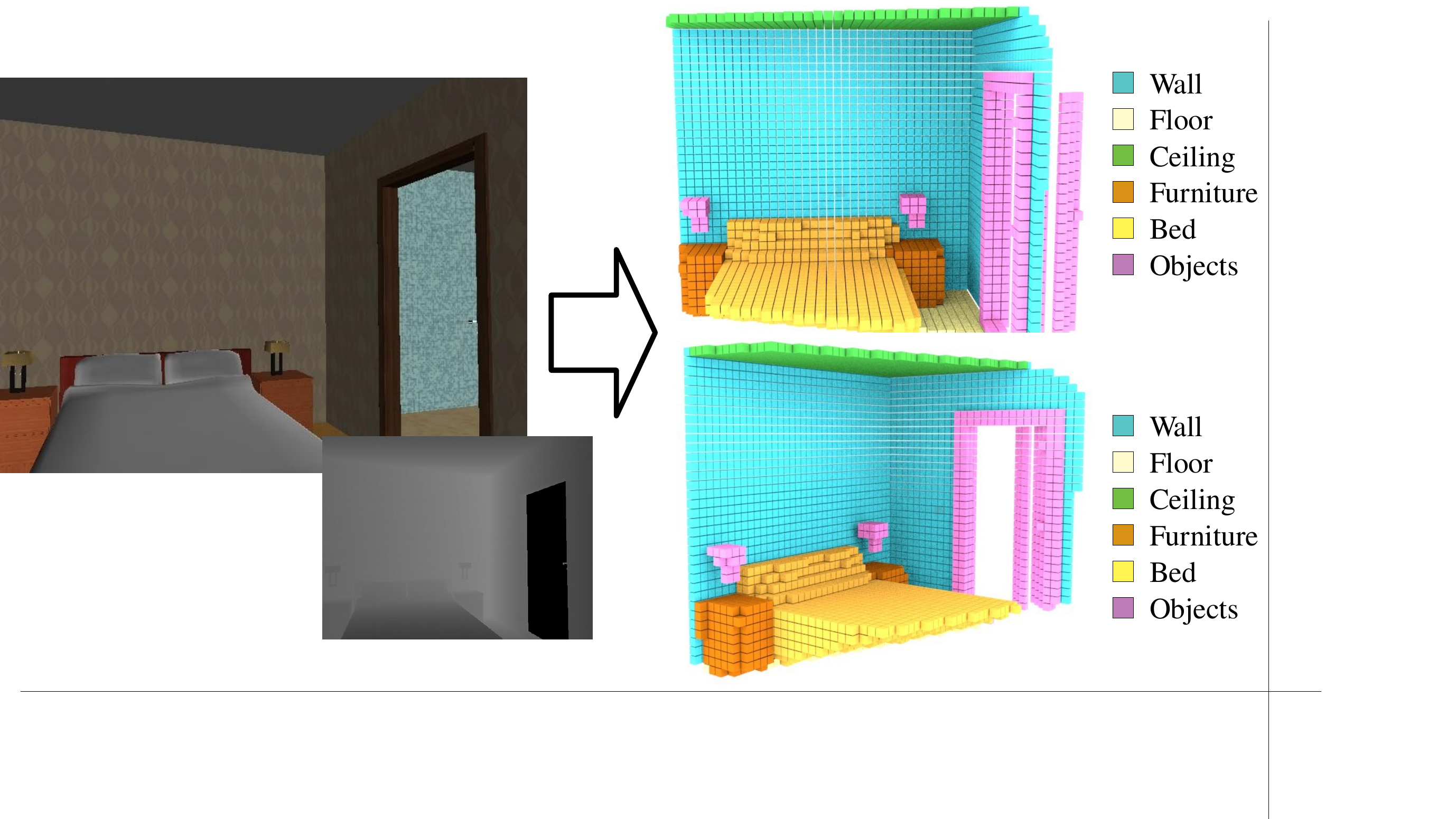}
\end{center}
   \caption{Semantic scene completion. Given an RGB-D image, the goal is to infer a complete 3D occupancy grid with associated semantic labels.}
\label{fig:1}
\end{figure}

Currently available  low  cost  RGB-D  sensors generate data form a single viewing position
and cannot handle occlusion among objects in the scene. 
For instance, in the scene depicted on the left part of Figure \ref{fig:1}, parts of
the wall, floor and furniture are occluded by the bed.
There is also self-occlusion: the interior of the bed, its sides and
its rear surfaces are hidden by the visible surface.

Given a partial 3D scene model acquired from a single RGB-D image, the goal of scene completion is to generate a complete 3D 
volumetric representation where each voxel is labelled as
occupied by some object or free space. 
For occupied voxels, the goal of {\em semantic} scene completion
is to assign a label that indicates to which class of object
it belongs, as illustrated on the right part of Figure \ref{fig:1}. 

Before 2018, most of the work on scene reasoning only partially addressees this problem. A number of approaches only infer labels of the visible surfaces \cite{Gupta_2013,Qi_2017,ren_2012}, while others only
consider completing the occluded part of the scene, without semantic
labelling \cite{firman_2016}. Another line of work focuses on single
objects, without the scene context~\cite{Nguyen_2016}.

The term semantic scene completion was introduced by Song \etal \cite{song_semantic_2017},
who showed that scene completion and semantic labelling are intertwined and training a CNN to jointly deals with both tasks can lead to better results. 
Their approach only uses depth information, ignoring all information from RGB channels. 
Colour information is expected to be
useful to distinguish objects that approximately share the same plane in the 3D space, and thus, are hard to be distinguished using only depth.
Examples of such instances are flat objects attached to the wall, such as posters, paintings and flat TVs. Some types of closed doors and windows are also problematic for depth-only approaches.

Recent research also explored colour information from on RGB-D images to improve semantic scene completion scores. Some methods project colour information to 3D in a naive way, leading to a problem of data sparsity in the voxelised data that is fed to the 3D CNN  \cite{guedes_semantic_2018}, while others uses RGB information to train a 2D segmentation network and then project generated features to 3D, requiring a complex two step training process \cite{garbade_two_2018, See_and_think_2018}.

Our work focuses on enhancing semantic scene segmentation scores using information from both depth and colour of RGB-D images in an end-to-end manner. In order to address the RGB data sparsity issue, we introduce a new strategy for encoding information extracted from RGB image in 3D space. We also present a new end-to-end 3D CNN architecture to combine and represent the features from colour and depth. Comprehensive experiments are conducted to evaluate the main aspects of the proposed solution. Results show that our fusion approach can enhance results of depth-only solutions and that EdgeNet achieves equivalent performance to current state-of-the-art fusion approach, with a much simpler training protocol.

To summarise, our main contributions are:
\begin{itemize}
\itemsep0em 
\item EdgeNet, a new end-to-end CNN architecture that fuses depth, RGB edge information to achieve state-of-the-art performance in semantic scene completion with a much simpler approach;
\item a new 3D volumetric edge representation using flipped signed-distance functions which improves performance and unifies data agregation for semantic scene completion from RGBD;
\item a more efficient end-to-end training pipeline for semantic scene completion with relation to previous approaches.
\end{itemize}

\section{Related Work}

Scene semantic completion in 3D is a problem that was introduced relatively recently.
Previous approaches to 3D SSC rely on Fully Convolutional Neural Network architectures (FCNs, introduced in \cite{FCNN}) and use SUNCG 
and NYUDv2 
as training sources (these datasets are described in Section~\ref{sec:datasets}).
We classify approaches into three main groups, based on the type of input of the semantic completion CNN: depth maps only, depth maps plus RGB and depth maps plus 2D segmentation maps.

\subsection{Depth maps only} 
 Song \etal\cite{song_semantic_2017} used depth maps from the SUNCG synthetic dataset to train a typical contracting fully convolutional CNN with 3D dilated convolutions, called SSCNet. They showed that jointly training for segmentation and completion leads to better results, as both tasks are inherently intertwined. To deal with data sparsity after projecting depth maps from 2D to 3D, the authors used a variation of Truncated Signed Distance Function (TSDF) that they called Flipped TSDF (F-TSDF). 
 Zhang \etal\cite{zhang_semantic_2018} used dense conditional random field to enhance SSCNet results. 
 Guo and Tong \cite{guo_view-volume_2018} applied a sequence of 2D convolutions to the depth maps, used a projection layer to projected the features to 3D and feed the output to a 3D CNN. 
 
 All solutions in this category are end-to-end approaches, in other words, the network is trained as a whole, with no need for extra training stages for specific parts. EdgeNet is an end-to-end network as well. RGB edges are aggregated in the same training pipeline of the depth information. 

\subsection{Depth maps plus RGB} 

Guedes \etal\cite{guedes_semantic_2018} reported preliminary results obtained by adding colour to an SSCNet-like architecture. In addition to the F-TSDF encoded depth volume, they used three extra projected volumes, corresponding to the channels of the RGB image, with no encoding, resulting in 3 sparse volumetric representation of the partially observed surfaces. The authors reported no significant improvement using the colour information in this sparse manner.

\subsection{Depth maps plus 2D segmentation}

Models in this category use a two step training protocol, where a 2D segmentation CNN is first trained and then it is used to generate input to a 3D semantic scene completion CNN. Current models differ in the way the generated 2D information is fed into the 3D CNN.  

Garbade \etal \cite{garbade_two_2018} used a pre-trained 2D segmentation CNN with a fully connected CRF \cite{CRF} to generate a segmentation map, which, after post-processing, was projected to 3D. 
Liu \etal \cite{See_and_think_2018} used depth maps and RGB information as input to an encoder-decoder 2D segmentation CNN. The encoder branch of the 2D CNN is a ResNet-101 \cite{ResNet}  and the decoder branch contains a series of dense upsampling
convolutions. The generated features from the 2D CNN are then reprojected to 3D using camera parameters, before being fed into a 3D CNN. The authors showed results using 2 different strategies to fuse depth and RGB: SNetFusion performs fusion just after the 2D segmentation network, while TNetFusion only performs fusion after the 3D convolutional network. TNetFusion achieves higher performance, with a much higher computational cost.  The 2D CNN is also pre-trained offline.  

Using 2D segmentation maps on 3D SSC brings an additional complexity to the training phase which is trains and evaluates the 2D segmentation network prior to the 3D CNN training. 
In this work, we propose an end-to-end approach to fuse information from depth and colour, where the network can be trained and evaluated as a whole, and still achieves state-of-the-art performance.

\section{Our solution: EdgeNet}

Our proposed solution is the first end-to-end approach that successfully uses information from RGB to improve semantic scene completion performance over depth only. It consists in a novel approach to encode information from RGB edges and depth maps and a new 3D CNN architecture to fuse both modalities that we call EdgeNet.

\subsection{Encoding edges in 3D}
\label{sec:f-tsdf}
As discussed earlier, colour information should complement depth maps for 3D semantic scene completion. However, combination of these modalities in a meaningful representation for learning is not trivial. Guedes \etal\cite{guedes_semantic_2018} naively added 3 channels to each voxel to insert R, G and B colour information into the representation, with no encoding. In this way, the vast majority of voxels have no colour data while only those at the visible surface have a colour value. This explains why they do not improve on the previous approach using depth only.
Song \etal \cite{song_semantic_2017} demonstrate that F-TSDF encoding plays an important role in feeding a projected depth map to a 3D CNN and produces better results than TSDF and other encoding techniques.

Given a sparse 3D voxel volume, the Truncated Signed Distance Function (TSDF) consists in computing the Euclidean distance of each empty voxel to the nearest occupied voxel. The signal of occluded regions is set to be negative, while visible regions are given positive values. Near the occupied surface, TSDF produces a value that tends to zero on both sides (and its first derivative tends to zero as well). TSDF values are normalised to [-1,1]. Flipped TSDF (F-TSDF) follows the same principle, but the absolute values of both visible and occluded regions are flipped: 
\begin{equation}
\operatorname{F-TSDF} = \text{sign}(\operatorname{TSDF}) \cdot (1 - |\operatorname{TSDF}|) .
\label{eq:3}
\end{equation}
A discontinuity near the occupied surface (from -1 to 1) occurs and the first derivative tends to infinity.

F-TSDF encoding of volumetric data can be easily applied to depth maps after 3D projection because each voxel carries binary information: occupied or free. On the other hand, F-TSDF can not straightforwardly be applied to RGB or semantic segmentation maps, because they are not binary.

To deal with this problem, we introduce a new strategy to fuse colour appearance and depth information for 3D semantic scene completion. 
Our approach exploits edge detection in the image, which gives a 2D binary representation of the scene that can highlight objects that are hard to detect in depth maps. For instance, a poster on a wall is expected to be invisible in a depth map, especially after down-sampling. On the other hand, RGB edges highlight the presence of that object.

The main advantage of extracting edges and projecting them to 3D is the possibility to apply F-TSDF on both edges and surface volumes, as they are both binary, providing two meaningful input signals to the 3D CNNs.
Another advantage is that due to their simplicity,
edges are more transferable, removing the need for
the application of a domain adaptation method when learning from synthetic images and applying on real images.

\begin{figure}[t]
\centering  
\begin{subfigure}{0.20\textwidth}
\includegraphics[width=\linewidth]{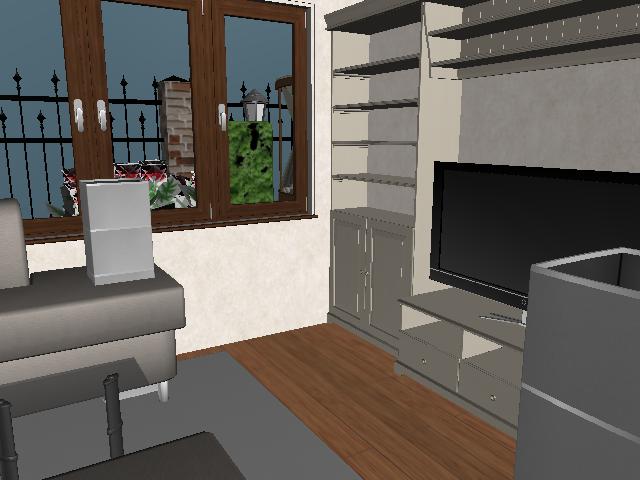}
\caption{}
\label{fig:depth3d_original}
\end{subfigure}
\begin{subfigure}{0.20\textwidth}
\includegraphics[trim=200 58 200 58, clip, width=\linewidth]{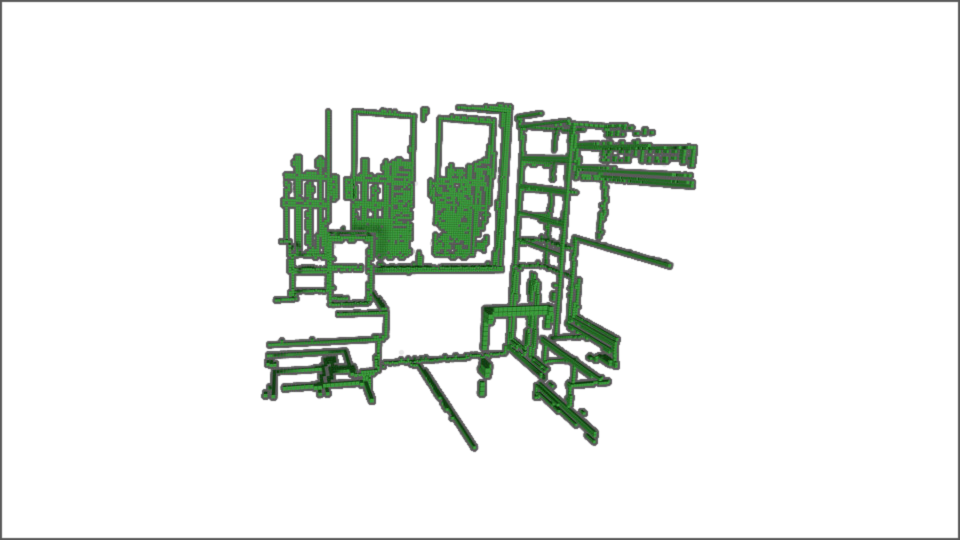}
\caption{}
\label{fig:depth3d_final}
\end{subfigure}
\caption{Projection of Edges to 3D: 
(a) original RGB image, 
(b) voxelized edges after projection.}
\label{fig:edges3d}
\end{figure}

\begin{figure}[t]
\centering  
\begin{subfigure}{0.18\textwidth}
\includegraphics[width=\linewidth]{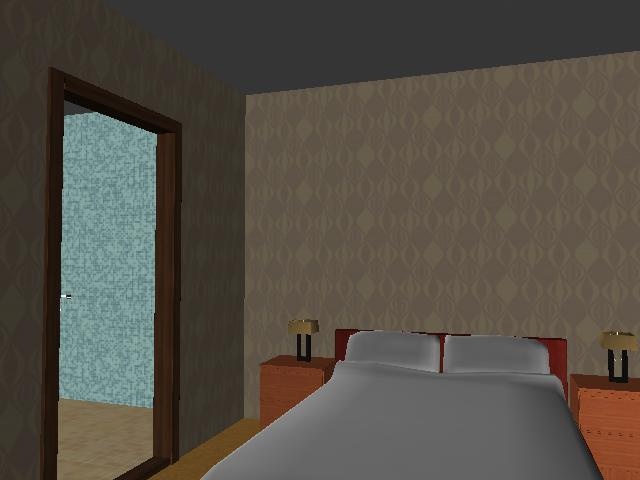}
\caption{\label{fig:edges-room_original}}
\end{subfigure}
\begin{subfigure}{0.28\textwidth}
\includegraphics[trim=0 150 0 50, clip, width=\linewidth]{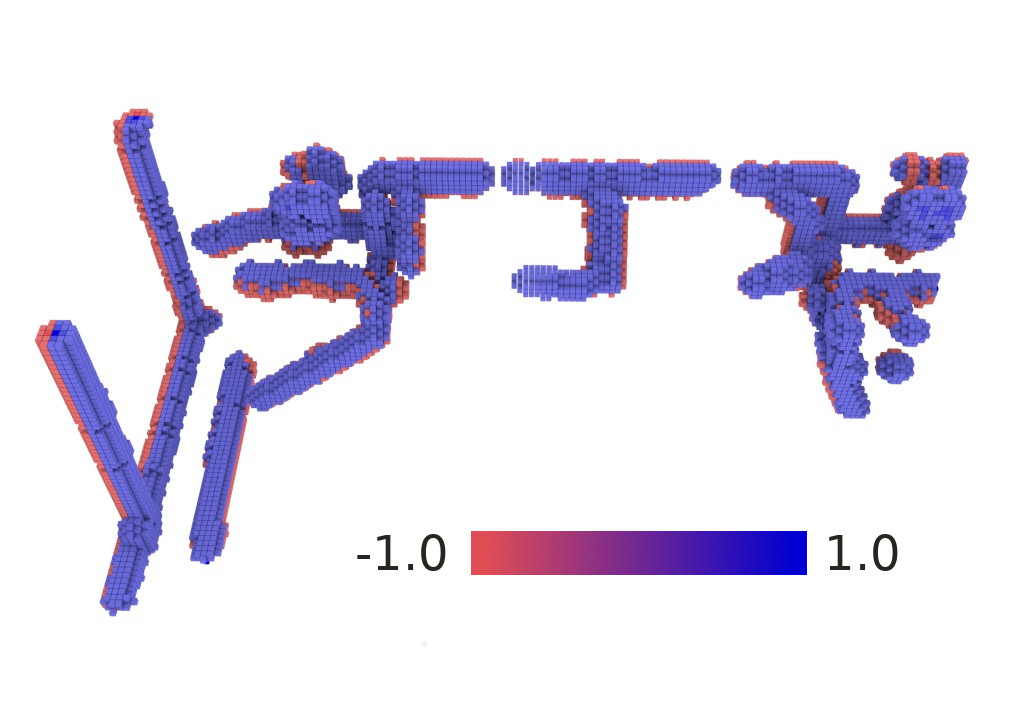}
\caption{\label{fig:edges-room_ftsdf}}
\label{fig:edges-room-legend}
\end{subfigure}
\caption{
(a) original scene. 
(b) F-TSDF of edges in 3D. The edge image is a horizontal cut of the scene, taken just above the bed. Only F-TSDF values with absolute value greater than 0.8 are shown (best viewed in colour).}
\label{fig:tsdf_edges}
\end{figure}

We apply F-TSDF to 3D edges, similarly to F-TSDF applied to 3D surfaces: for each voxel in the edge volume, we look for the nearest edge to calculate the Euclidean distance. Visible and occluded voxels are related only to edges, not to surfaces. 
We use the standard Canny edge detector \cite{canny_edge_detection} to perform edge detection. Each edge location is projected to a point in the 3D space using its depth information and the camera calibration matrix. The resulting point cloud is voxelised in the same way as the depth point cloud, resulting in a sparse volume of $240\times144\times240$ voxels. Figure \ref{fig:edges3d} shows a scene from the SUNCG dataset and its corresponding edges projected to 3D. 
Figure \ref{fig:edges-room_ftsdf} shows in detail a region of the projected edges of \ref{fig:edges-room_original} after F-TSDF encoding. Note that the greatest gradients occur along the edges.

\begin{figure*}
\centering  
\includegraphics[trim=0 0 0 0, clip,width=\linewidth] {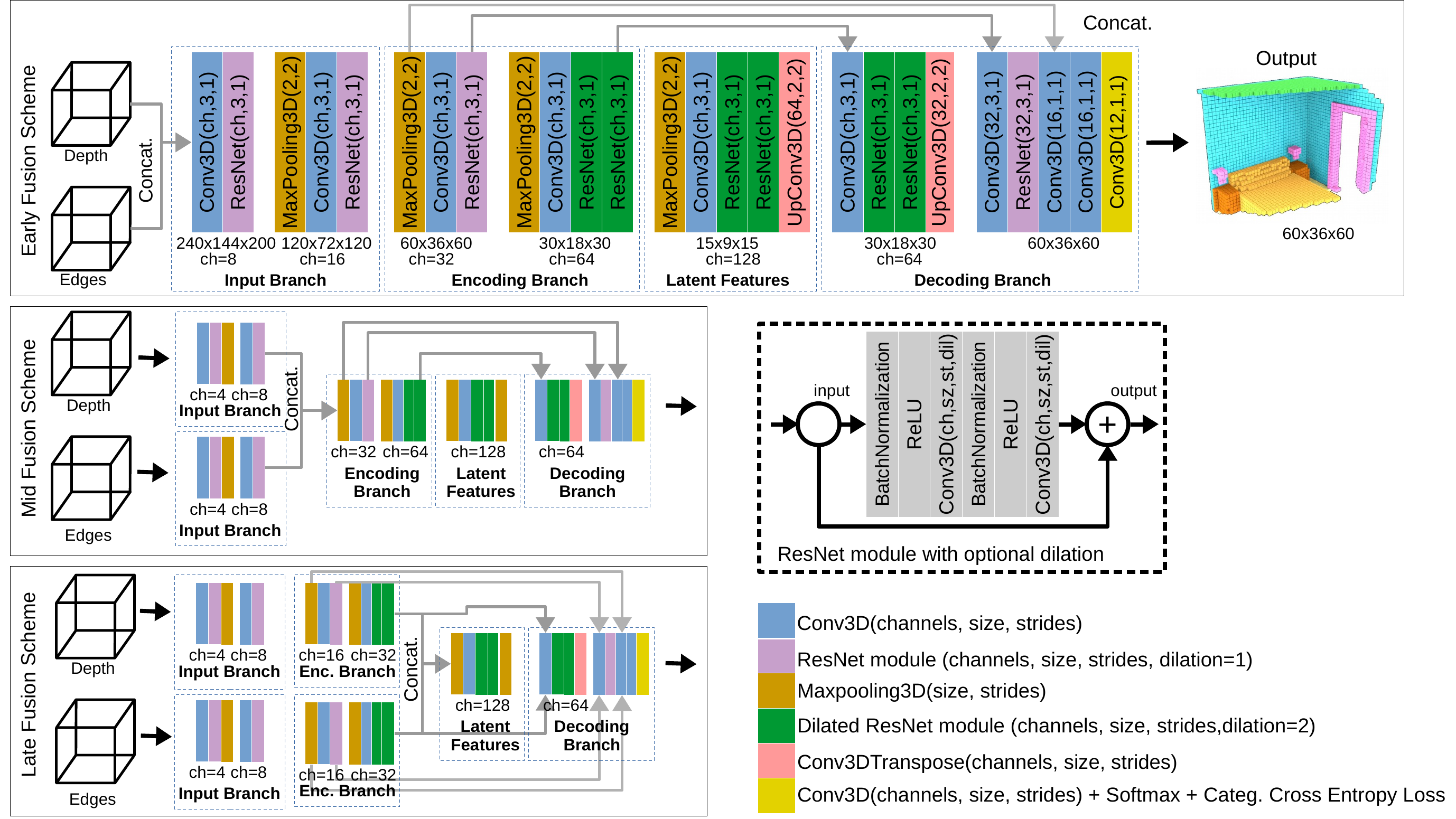}
\caption{EdgeNet architecture and fusion schemes (best viewed in colour).}
\label{fig:arch}
\end{figure*}

\subsection{EdgeNet architecture}

In order to combine depth and edge modalities, we propose a new 3D semantic segmentation CNN architecture that we call \textbf{EdgeNet}. Our proposed solution is a 3D CNN inspired by the U-Net design \cite{unet_2015} which has been successfully used in many 2D semantic segmentation problems, and is presented in Figure \ref{fig:arch}. We address the degradation problem of deeper networks \cite{He_2015_CVPR}, by replacing simple convolutional blocks of U-Net by ResNet modules \cite{ResNet}. On lower resolutions, the ResNet modules uses dilated convolutions  to improve the receptive field. To match the resolution of the output, the input branch reduces the resolution to 1/4 of the input. Next blocks follows encoder-decoder design and the last stage of the decoding branch is responsible for reducing the number of channels to match the desired number of output classes and loss calculations. 

\textbf{Depth and Edges Fusion Schemes.}
The encoder-decoder structure of EdgeNet allows us to evaluate three fusion schemes: Early Fusion (EdgeNet-EF), Middle Fusion (EdgeNet-MF) and Late Fusion (EdgeNet-LF). In EdgeNet-EF, just after F-TSDF encoding, both input volumes are concatenated and fed into the main network. In EdgeNet-MF, the input branch is divided into two parts while in EdgeNet-LF, both input and encoding branches are divided. To keep the same memory requirement in all fusion schemes, the total quantity of channels in all scheme is always the same.

\textbf{Data balancing and loss function.}
In volumetric data, occluded and occupied voxels are highly unbalanced, so we use a weighted version of categorical cross entropy as the loss function to train our models. To obtain the weights, for each training batch, we randomly initialize a tensor $rand_{occl}$ of the same shape as the batch with ones and zeroes using the ratio $r=(2\sum{occu}/\sum{occl})$, where $occl$ and $occu$ are two tensors obtained from the previously calculated occupancy grid relative to occluded and occupied voxels. The final weight tensor is $w=occu + occl \odot rand_{occl}$, where $\odot$ denotes the Hadamard product. 
Let $p$ be the predicted probabilities of the 12 classes for each voxel and $y$ be the one hot encoded ground truth tensor. The categorical cross entropy loss function is then given by 
\begin{equation}
L_{cce}(p,y) = -\sum{(w \odot y \odot \log{p})} .
\label{eq:4}
\end{equation}

\subsection{Training pipeline with offline data preparation}

 As F-TSDF calculation is computationally intensive, to reduce overall training time, the F-TSDF volumes that feed the models are preprocessed off-line once. The preprocessed dataset is then stored, and may be used as many times as needed, including by different models. Following previous works, we rotate the 3D Scene to align it with gravity and have room orientation based on the Manhattan assumption. We fixed the dimensions of the 3D space to 4.8 m horizontally, 2.88 m vertically and 4.8 m in depth. Voxel grid size is 0.02 m, resulting in a $240\times144\times240$ 3D volume.  The TSDF truncation value is 0.24 m. Surface and edge projection as well as F-TSDF encoding of all volumes are done in this stage.
During preprocessing, we also calculate an occupancy grid where we distinguish occupied voxels inside the room and FOV; non-occupied occluded voxels inside the room and FOV; and all other voxels. This occupancy grid will be further used to balance the dataset during training time.

\section{Experiments}
In this section we describe the datasets and the evaluation protocol used in this paper. 

\subsection{Datasets}
\label{sec:datasets}

We train and validate our proposed approach on SUNCG \cite{song_semantic_2017} and NYUDv2 \cite{silberman_2012} datasets. 
The SUNCG dataset consists of about 45K synthetic scenes from which more than 130K 3D scenes were rendered with corresponding depth maps and ground truth, divided in train and test datasets. As the original training and test sets did not include RGB images, we extracted the camera poses from the provided ground truth and rendered a new set of depth and RGB images from the SUNCG synthetic scenes. To avoid misalignments, the ground truth volumes were regenerated from the scene meshes.  

NYUDv2 is a widely used dataset of indoor scenes that includes depth and RGB images captured by the Kinect depth sensor, divided in 795 samples for training and 654 for test. Following the majority of works in semantic segmentation we used ground truth obtained by voxelizing the 3D mesh annotations from \cite{Guo2015} and and mapped object categories based on \cite{Handa2015}.

\subsection{Training protocols}
Our experiments consists in training our models from scratch on SUNCG and NYUDv2, and also fine-tuning models trained from SUNCG to NYUDv2. 
For experiments in which we trained our models from scratch, we use the technique known as One Cycle Learning \cite{Smith_2018}, which is a combination of Curriculum Learning \cite{Bengio_2009} and  Simulated Annealing \cite{Aarts:1989:SAB:61990}. After some preliminary tests, we found 0.01 to be a good base learning rate.  We use a maximum of 30 epochs, in order to maintain total training time in a acceptable limit. Following \cite{Smith_2018}, we start with the base learning rate and linearly increase the effective learning until 0.1 in the 10th epoch, then  linearly decrease the learning rate until reach the start-up level in the 20th epoch. During the annealing phase, we linearly go from 0.01 to 0.0005 in a further 10 epochs.
Due to GPU memory size constraints, we use a batches of 3 samples. We also use SGD optimizer with a momentum of 0.9 and decay of 0.0005 in all experiments, as used in most previous works. For SUNCG, each epoch consists of 30,540 scenes randomly selected from the whole training set. For NYUDv2, each epoch comprises the whole training set. 
For fine tuning, we initialize the network with parameters trained on SUNCG and use the standard training policy with SGD with fixed learning rate of 0.01 and 0.0005 of weight decay. 

Thanks to our lightweight training pipeline with offline F-TSDF preprocessing, our training time is only 4 days on SUNCG and 6 hours on NYUDv2, using a GTX 1080 TI.

\subsection{Evaluation}

For the semantic scene completion task, we report the Intersection over Union (IoU) of each object class on both the observed and occluded voxels. For the scene completion task, all non-empty object classes are considered as one category, and we report Precision, Recall and IoU of the binary predictions on occluded voxels. Voxels outside the view or the room are not considered.

\begingroup
\setlength{\tabcolsep}{2pt} 
\renewcommand{\arraystretch}{1} 

\begin{table*}
  \centering
  \begin{tabular}
      { c | c | c c c | c c c c c c c c c c c c}
      \hline
     \multirow{ 2}{*}{input}&\multirow{ 2}{*}{model} & \multicolumn{3}{c|}{scene completion} & \multicolumn{12}{c}{semantic scene completion (IoU, in percentages)} \\
   
    & & prec. & rec. & IoU & ceil. & floor & wall & win. & chair & bed & sofa & table & tvs & furn. & objs. & avg. \\
    \hline
    \multirow{ 5}{*}{d}&SSCNet\cite{song_semantic_2017} & 76.3 & \textbf{95.2} & 73.5 & 96.3 & 84.9 & 56.8 & 28.2 & 21.3 & 56.0 & 52.7 & 33.7 & 10.9 & 44.3 & 25.4 & 46.4\\
   &SSCNet*& 92.7&89.7&83.8&97.0&94.6&74.3&51.1&43.7&78.2&70.9&49.5&45.2&61.0&51.3&65.2\\

  &DCRF \cite{zhang_semantic_2018}& – & – & –& 95.4 &  84.3 &  57.7 & 24.5 &  28.2 &  63.4 & 55.3 &  34.5 &  19.6 &  45.8 &  28.7 & 48.8 \\
  &VVNetR-120 \cite{guo_view-volume_2018}& 90.8 & 91.7 & 84.0 & \textbf{98.4} &  87.0 &  61.0 & 54.8 &  49.3 &  83.0 & \textbf{75.5} &  55.1 &  43.5 &  68.8 &  57.7 & 66.7 \\
  &EdgeNet-D & 93.1&90.4&84.8&97.2&94.4&78.4&56.1&50.4&80.5&73.8&54.5&49.8&69.5&59.2&69.5\\
    \hline
    \multirow{ 2}{*}{d+s}&SNetFuse\cite{See_and_think_2018} & 56.7 & 91.7 & 53.9 & 65.5 & 60.7 & 50.3 & 56.4 & 26.1 & 47.3 & 43.7 & 30.6 & 37.2 & 44.9 & 30.0 & 44.8\\
   &TNetFuse\cite{See_and_think_2018} & 53.9 & 95.2 &52.6 &60.6&57.3&53.2&52.7&27.4&46.8&53.3&28.6 &41.1&44.1 &29.0& 44.9\\

      \hline
      
   \multirow{4}{*}{d+e}
   &SSCNet-E & 92.8&89.6&83.8&97.0&94.5&74.6&51.8&43.9&77.0&70.8&49.3&49.2&62.1&52.0&65.7\\
   &EdgeNet-EF(Ours) &\textbf{93.7}&90.3&\textbf{85.1}&97.2&94.9&\textbf{78.6}&57.4&49.5&80.5&74.4&\textbf{55.8}&51.9&70.1&\textbf{62.5}&\textbf{70.3}\\
      &EdgeNet-MF(Ours) & 93.3 & 90.6 & \textbf{85.1} & 97.2 & \textbf{95.3} & 78.2 & \textbf{57.5} & \textbf{51.4} & \textbf{80.7} & 74.1 & 54.5 & \textbf{52.6} & \textbf{70.3} & 60.1 & 70.2\\
      &EdgeNet-LF(Ours) & 93.0 & 89.6 & 83.9 & 97.0 & 94.6 & 76.4 & 52.0 & 44.6 & 79.8 & 71.5 & 48.9 & 48.3 & 66.1 & 55.9 & 66.8\\
 \hline
    
  \end{tabular}
  \caption{\textbf{Results and ablation studies on SUNCG test set}. We took SSCNet as a baseline and show the effect of each one of the main aspects of our proposed approach. Column `input' indicates the type of input: d = depth only;  d+e = depth + edges. SSCNet* is our implementation of the original SSCNet, with our training pipeline. EdgeNet-D has the same architecture of the other versions of EdgeNet, but the edge volume is not fed into the network. EdgeNet-EF achieves the best overall scores and  surpassed VVNetR-120 by 3.3\% on average IoU for semantic scene completion.}
  \label{tab:2}
\end{table*}
\endgroup

\begingroup
\setlength{\tabcolsep}{2pt} 
\renewcommand{\arraystretch}{1} 

\begin{table*}
  \centering
  \begin{tabular}
      { c | c | c | c c c | c c c c c c c c c c c c}
  \hline
      \multirow{ 2}{*}{train}&\multirow{ 2}{*}{input}&\multirow{ 2}{*}{model} & \multicolumn{3}{c|}{scene completion} & \multicolumn{12}{c}{semantic scene completion (IoU, in percentages)} \\
       & & & prec. & rec. & IoU & ceil. & floor & wall & win. & chair & bed & sofa & table & tvs & furn. & objs. & avg. \\
  \midrule\midrule
       \multirow{4}{*}{SUNCG}&\multirow{1}{*}{d} &SSCNet\cite{song_semantic_2017} & 55.6 & 91.9 & 53.2 & 5.8 & 81.8 & 19.6 & 5.4 & 12.9 & 34.4 & 26 & 13.6 & 6.1 & 9.4 & 7.4 & 20.2\\ 
       \cline{2-18}
       &\multirow{3}{*}{d+e}
       &EdgeNet-EF(Ours) &  \textbf{61.9} &80.0&\textbf{53.6} &9.1&\textbf{92.9}&18.3&5.7&15.8&40.4&30.7&9.2&3.3&13.7&11.6&22.8\\
       &&EdgeNet-MF(Ours) & 60.7 & 80.3 & 52.8 & \textbf{11.0} & 92.3 & \textbf{20.5} & 7.2 & \textbf{16.3} &42.8 & \textbf{32.8} & \textbf{10.5} & \textbf{6.0} &\textbf{ 15.7} &\textbf{11.8} & \textbf{24.3}\\ 
       &&EdgeNet-LF(Ours) & 59.9 & \textbf{80.5} & 52.3 & 3.2 & 87.1 & 19.9 & \textbf{8.6} & 15.4 & \textbf{ 43.5} & 32.3 & 8.8 & 4.3 & 13.7 &10.0 & 22.4\\ 
  \midrule\midrule
      \multirow{4}{*}{NYU}&\multirow{1}{*}{d}&SSCNet\cite{song_semantic_2017} 
      & 57.0 & \textbf{94.5} & 55.1 & 15.1 & 94.7 & 24.4 & 0.0 & \textbf{12.6} & 32.1 & 35.0 & \textbf{13.0} & \textbf{7.8} & 27.1 & 10.1 & 24.7\\  
     \cline{2-18}
      &\multirow{ 3}{*}{d+e}
       &EdgeNet-EF(Ours) &  \textbf{78.1}  & 65.1 & 55.1 & \textbf{21.8} & \textbf{95.0} & 27.3 & \textbf{8.4} & 6.8 & \textbf{53.1} & 38.6 & 7.5 & 0.0 & 30.4 & \textbf{13.3} & 27.5\\
       &&EdgeNet-MF(Ours) & 76.0 & 68.3 & \textbf{56.1} & 17.9 & 94.0 & \textbf{27.8} & 2.1 & 9.5 &51.8 & \textbf{44.3} & 9.4 & 3.6 &\textbf{ 32.5} &12.7 & \textbf{27.8}\\ 
       &&EdgeNet-LF(Ours) & 75.5 & 67.5 & 55.4 & 19.8 & 94.9 & 24.4 & 5.7 & 7.2 & 50.3 & 38.8 & 10.0 & 0.0 &33.2 &12.2 & 27.0\\ 
   
  \midrule\midrule
      \multirow{10}{*}{\makecell{SUNCG \\ +\\NYU}} &\multirow{3}{*}{d}&SSCNet\cite{song_semantic_2017} & 59.3 & 92.9 & 56.6 & 15.1 & 94.6 & 24.7 & 10.8 & 17.3 & 53.2 & 45.9 & 15.9 & 13.9 & 31.1 & 12.6 & 30.5\\ 
      &&DCRF\cite{zhang_semantic_2018} & - & -& - &18.1 &  92.6 &  27.1 & 10.8 &  18.8 &  54.3 & 47.9 &  17.1 &  15.1 &  34.7 &  13.0 & 31.8\\ 
      &&VVNetR-120\cite{guo_view-volume_2018} & 69.8 &83.1& 61.1 & 19.3 & 94.8 & 28.0 & 12.2 & 19.6 & 57.0 & 50.5 & 17.6 & 11.9 & 35.6 & 15.3 & 32.9\\ 
      \cline{2-18}
      &\multirow{1}{*}{d+c}&Guedes \etal\cite{guedes_semantic_2018} & - & - & 56.6 & - & - & - & - & - & - & - & - & - & - & - & 30.5\\
      \cline{2-18}
      &\multirow{3}{*}{d+s}&Garbade \etal*\cite{garbade_two_2018} & 69.5 & 82.7 & \textbf{60.7} & 12.9 & 92.5 & 25.3 & 20.1 & 16.1 & 56.3 & 43.4 & 17.2 & 10.4 & 33.0 & 14.3 & 31.0 \\ 
      &&SNetFuse\cite{See_and_think_2018} & 67.6 & \textbf{85.9} & \textbf{60.7} & 22.2 & 91.0 & 28.6 & \textbf{18.2} & 19.2 & 56.2 & 51.2 & 16.2 & 12.2 & 37.0 & 17.4 & 33.6\\
      &&TNetFuse\cite{See_and_think_2018} & 67.3 & 85.8 & \textbf{60.7} & 17.3 & 92.1 & 28.0 & 16.6 & 19.3 & \textbf{57.5} & \textbf{53.8} & \textbf{17.7} & \textbf{18.5} & \textbf{38.4} & \textbf{18.9} & \textbf{34.4}\\
      \cline{2-18}
      &\multirow{3}{*}{d+e}&EdgeNet-EF(Ours) &77.0&70.0&57.9& 16.3 & \textbf{95.0} &27.9 & 14.2 & 17.9 & 55.4 & 50.8 & 16.5 & 6.8 & 37.3 & 15.3 & 32.1\\
      &&EdgeNet-MF(Ours) & \textbf{79.1} & 66.6 & 56.7 & \textbf{22.4} & \textbf{95.0} & \textbf{29.7} & 15.5 & \textbf{20.9} & 54.1 & 53.0 & 15.6 & 14.9 & 35.0 & 14.8 & 33.7\\
      &&EdgeNet-LF(Ours) & 77.6 & 69.5 & 57.9 & 20.6 & 94.9 & 29.5 & 9.8 & 18.1 & 56.2 & 50.5 & 11.4 & 5.2 & 35.9 & 15.3 & 31.6\\
  \hline
    
  \end{tabular}
  \caption{\textbf{Semantic scene completion results on NYUDv2 test set}. Column input indicates the type of input: d=depth only; d+s=depth and segmentation maps; d+e=depth and edges. Column train indicates dataset used for training the models. SUNCG + NYU means trained on SUNCG and fine tuned on NYUDv2. }
  \label{tab:1}
\end{table*}
\endgroup

\subsection{Experimental results}

We compare our results to semantic scene completion approaches that use depth-only \cite{guo_view-volume_2018,song_semantic_2017,zhang_semantic_2018}, depth plus RGB \cite{guedes_semantic_2018} and depth plus 2D segmentation maps \cite{garbade_two_2018,See_and_think_2018}. We also investigate the effects of the main aspects of our proposed solution on SUNCG. Comparative results were extracted from the original papers.

\subsubsection {Ablation Studies and results on SUNCG}

 In Table \ref{tab:2}, investigate the effects of the main aspects of our proposed solution. At first, we analyse the effect of our training pipeline. We took SSCNet as a baseline and retrain it, using our light-weight training framework, that allows a batch size of 3 samples in comparison to the 1 sample batch size of original SSCNet. Results of that experiment are shown as SSNet*. We observed a large improvement on SSC scores just using our pipeline.

After isolating the effect of our training protocol, we investigate the effect of our encoder-decoder architecture, with dilated ResNet modules. To accomplish this, we used EdgeNet-D, that is the Ednet architecture fed only with depth, without edges. Once again we observed a high level of improvement, comparing to SSCNet*. EdgeNet-D also got the best overall scores amongst the depth-only approaches. Next experiment evaluates the effect of adding edges to an existing depth-only architecture. We took SSCNet and fed it with both depth and edges after F-TSDF encoding (SSCNet-E). We observed improvements compared to SSCNet* on overall scores and especially on hard-to-detect classes like TVs and objects.

Finally, we evaluate the benefits of adding Edges to our architeture in three fusion schemes: EdgeNet-EF, EdgeNet-MF and EdgeNet-LF.  Performance gains from EdgeNet-D show, once again, that adding edges is useful. A discussion about fusion schemes is provided on Section \ref{sec:discussion}. 

We also compare EdgeNet results to previous approaches. Overall, our proposed solutions achieve the best performance by a large margin. EdgeNet-EF achieves best average scores, while EdgeNet-MF achieves the best score in some classes. EdgeNet-EF surpassed VVNetR120, the best previous approach on average SSC, by 3.3\%. As expected, the highest improvements are observed on hard to detect classes, like objects and TVs. Although SUNCG is synthetic, evaluation on this dataset is quite important because of the poor quality of the ground truth in NYU, which impacts negatively accurate models like EdgeNet.

\subsection{Results on NYUDv2}

Table \ref{tab:1} shows the results of EdgeNet on NYUDv2 dataset and compares it with previous approaches. We compare results for models trained only on synthetic data, only on NYUDv2 and on both synthetic and NYUDv2 using fine tuning. 

On SUNCG-only and on NYUDv2-only training scenarios, EdgeNet-MF achieved the best overall scores on Scene Completion and Semantic Scene Completion. 
On SUNCG+NYU training scenario, however, TNetFuse presented the best 
result. EdgeNet-MF achieved best scores on structural elements and chair. It is worth mentioning that the NYUDv2 dataset  has severe ground truth errors and misalignment, so results are not precise, and small differences in results may be questioned (see Section \ref{sec:qualy}).

Despite these problems on NYU ground truth, EdgeNet achieves state-of-the-art level results with a much simpler and more computationally efficient training pipeline. EdgeNet is an end-to-end approach, and its memory consumption allows a batch size of 3 samples in a GTX 1080TI GPU, while TNetFuse requires a complex two step training procedure and uses a batch size of only 1 sample, in the same GPU.

\begin{figure*}[ht]
\centering  

\begin{subfigure}{.8\textwidth}
\includegraphics[trim=50 240 50 250, clip, width=\linewidth]{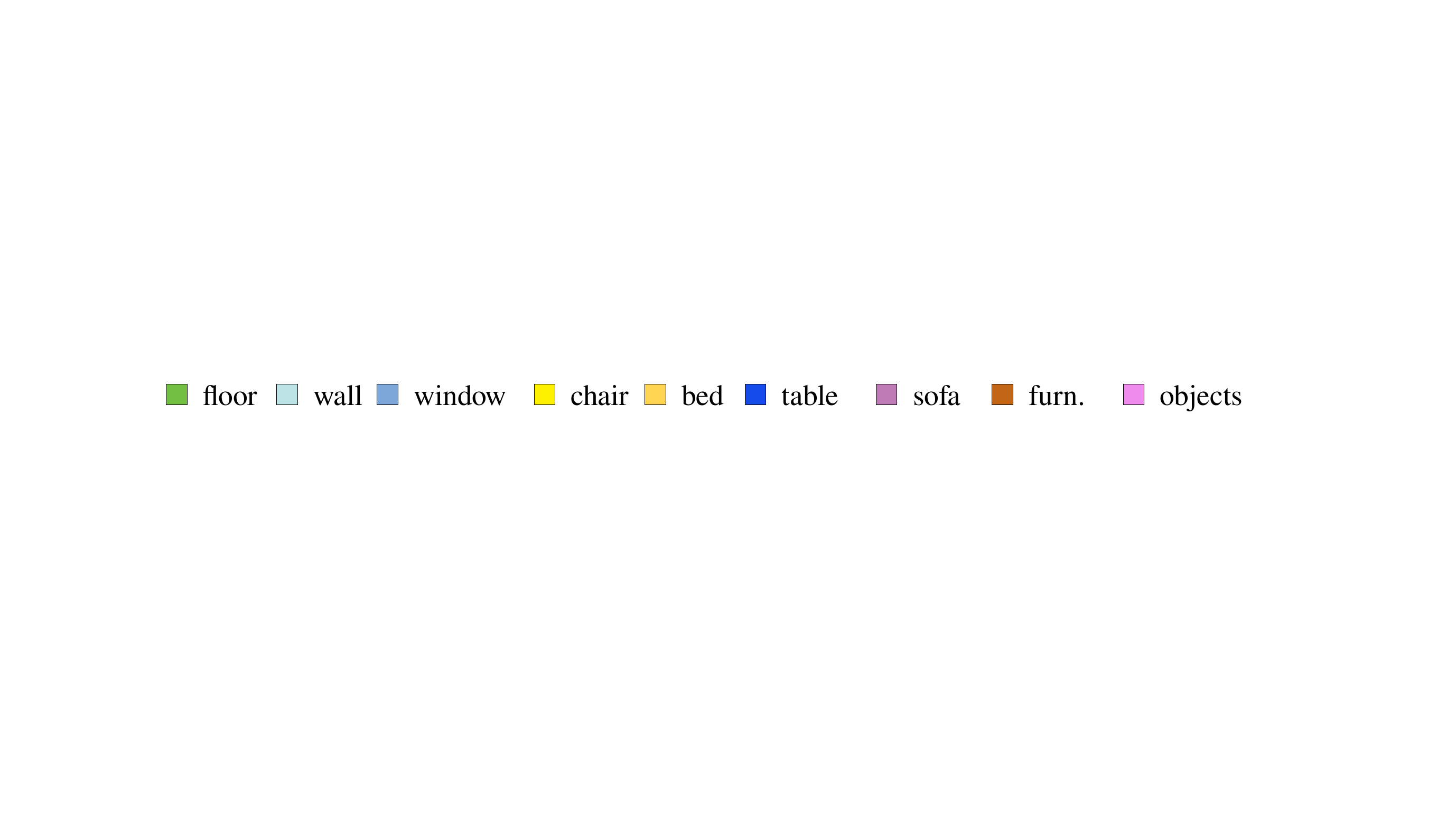}

\end{subfigure}

\bigskip


\begin{subfigure}{0.16\textwidth}
\includegraphics[width=\linewidth]{figures/NYU0524_0000_color.jpg}

\end{subfigure}
\begin{subfigure}{0.16\textwidth}
\includegraphics[trim=20 0 20 0, clip, width=\linewidth]{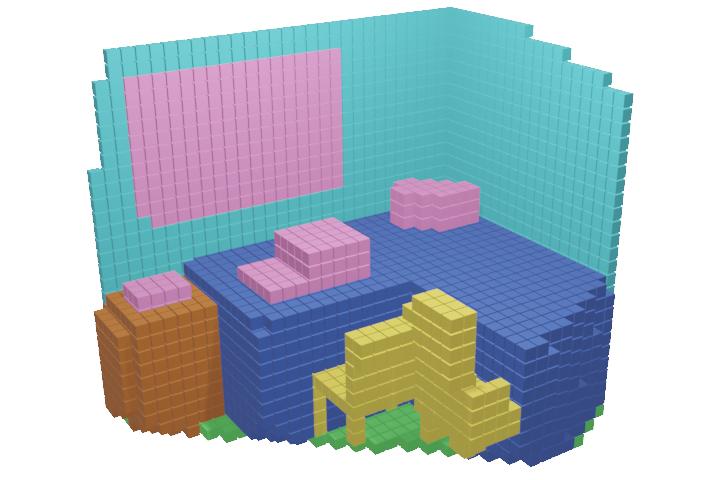}
\end{subfigure}
\begin{subfigure}{0.16\textwidth}
\includegraphics[trim=20 0 20 0, clip, width=\linewidth]{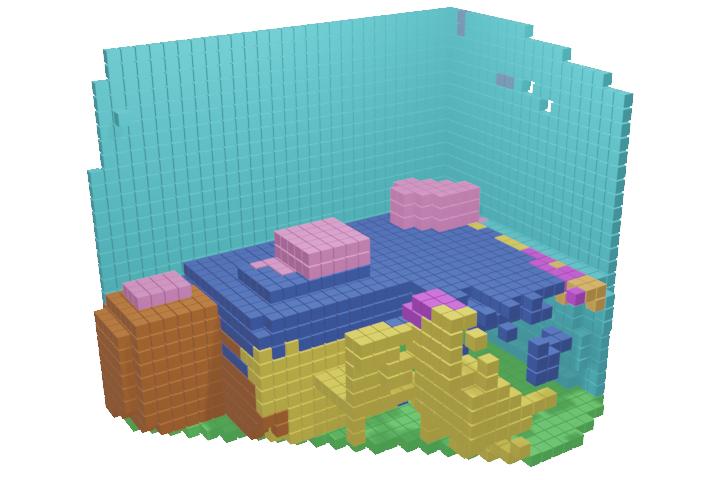}

\end{subfigure}
\begin{subfigure}{0.16\textwidth}
\includegraphics[trim=20 0 20 0, clip, width=\linewidth]{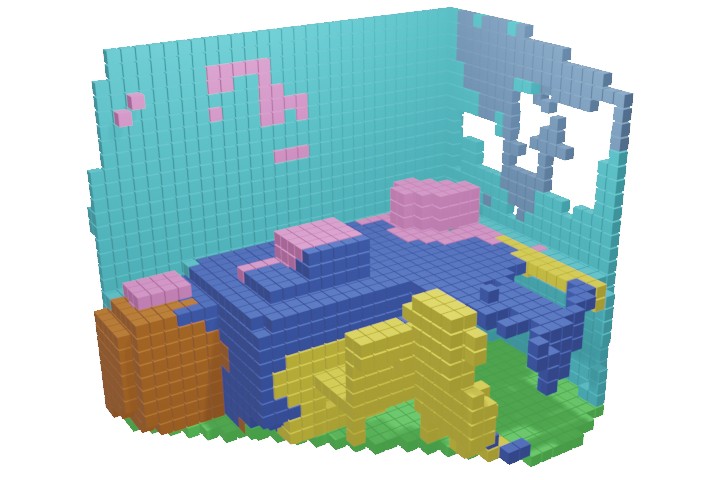}

\end{subfigure}
\begin{subfigure}{0.16\textwidth}
\includegraphics[trim=20 0 20 0, clip, width=\linewidth]{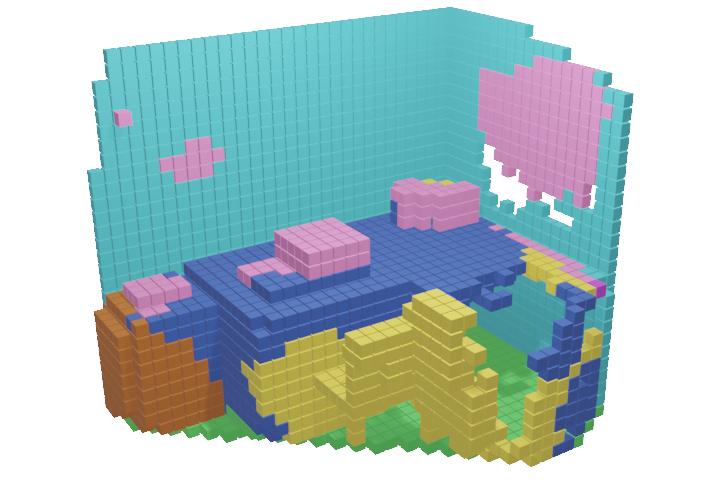}

\end{subfigure}
\begin{subfigure}{0.16\textwidth}
\includegraphics[trim=20 0 20 0, clip, width=\linewidth]{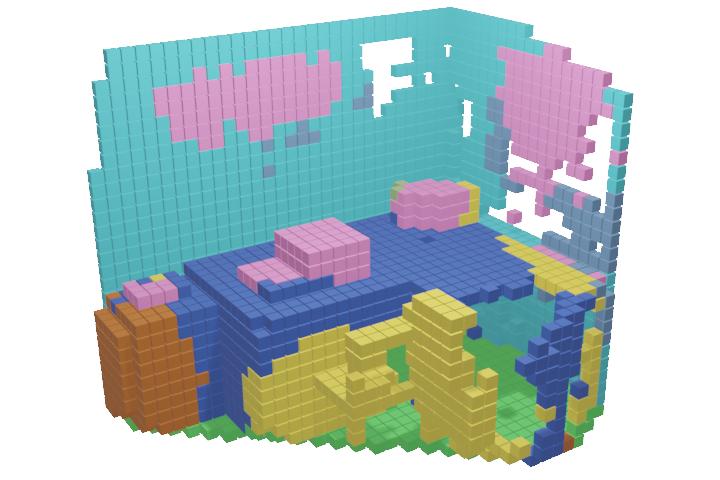}

\end{subfigure}


\begin{subfigure}{0.16\textwidth}
\includegraphics[width=\linewidth]{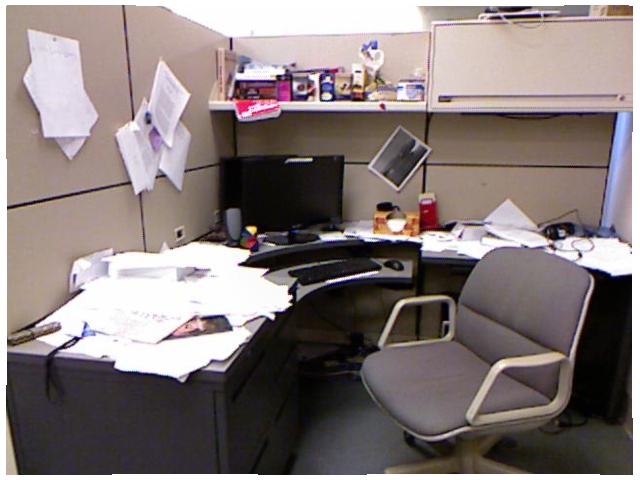}

\end{subfigure}
\begin{subfigure}{0.16\textwidth}
\includegraphics[trim=20 0 20 0, clip, width=\linewidth]{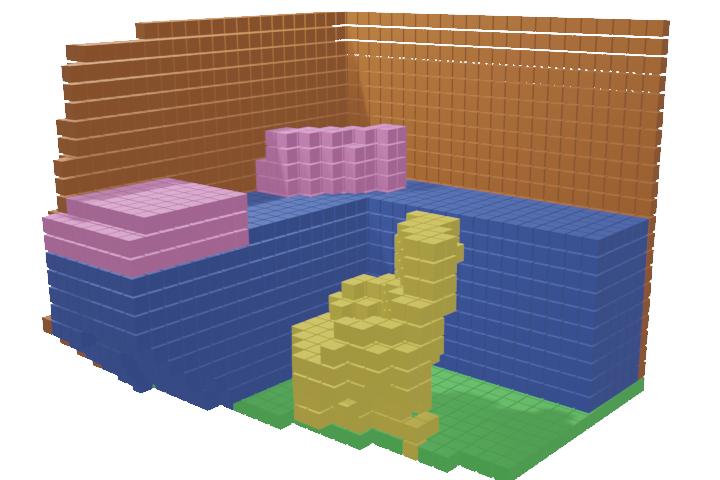}
\end{subfigure}
\begin{subfigure}{0.16\textwidth}
\includegraphics[trim=20 0 20 0, clip, width=\linewidth]{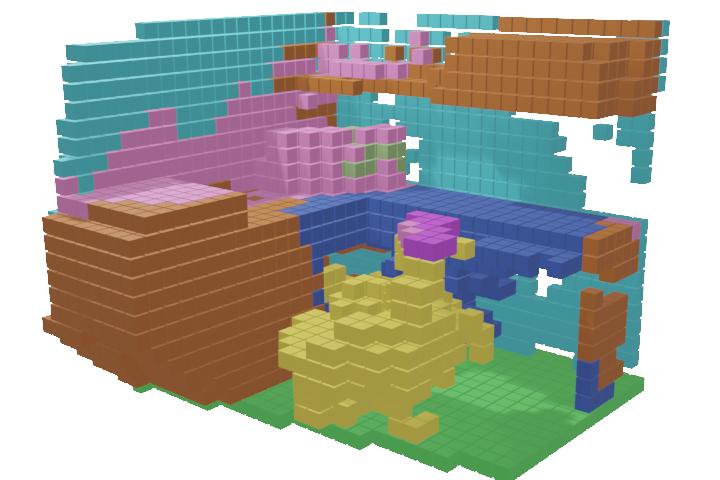}

\end{subfigure}
\begin{subfigure}{0.16\textwidth}
\includegraphics[trim=20 0 20 0, clip, width=\linewidth]{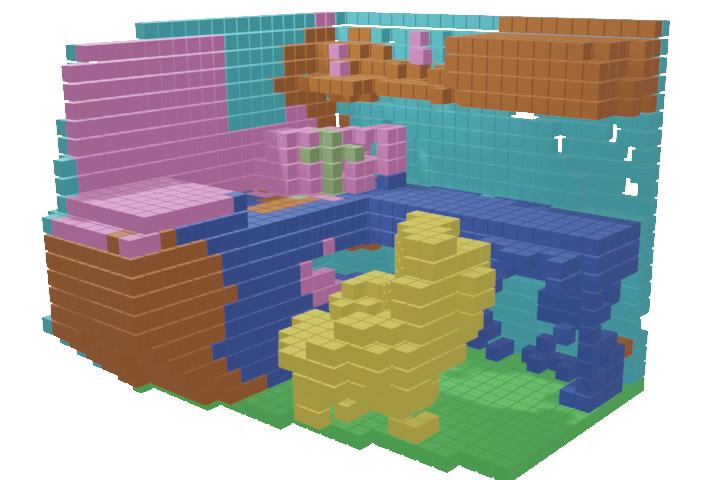}

\end{subfigure}
\begin{subfigure}{0.16\textwidth}
\includegraphics[trim=20 0 20 0, clip, width=\linewidth]{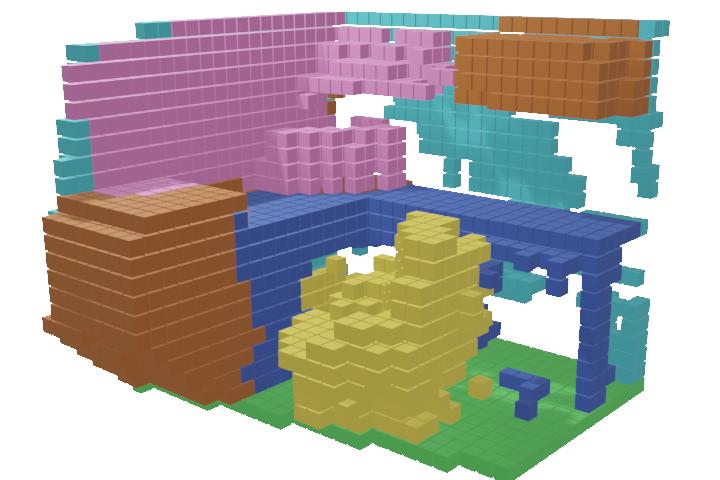}

\end{subfigure}
\begin{subfigure}{0.16\textwidth}
\includegraphics[trim=20 0 20 0, clip, width=\linewidth]{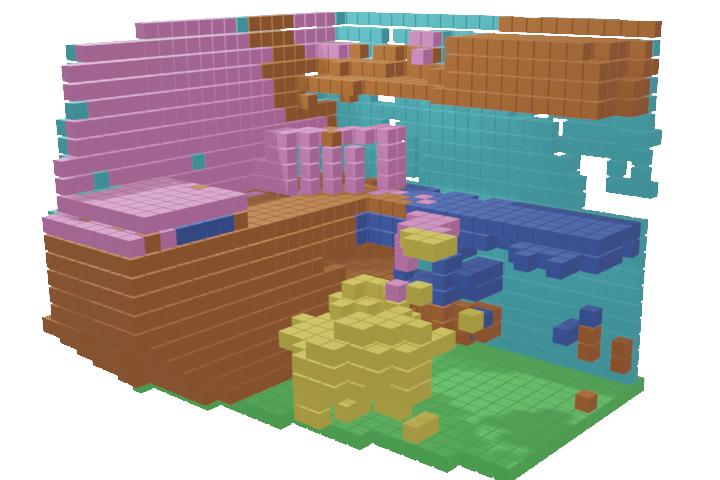}

\end{subfigure}


\begin{subfigure}{0.16\textwidth}
\includegraphics[width=\linewidth]{figures/NYU0959_0000_color.jpg}

\end{subfigure}
\begin{subfigure}{0.16\textwidth}
\includegraphics[trim=20 0 20 0, clip, width=\linewidth]{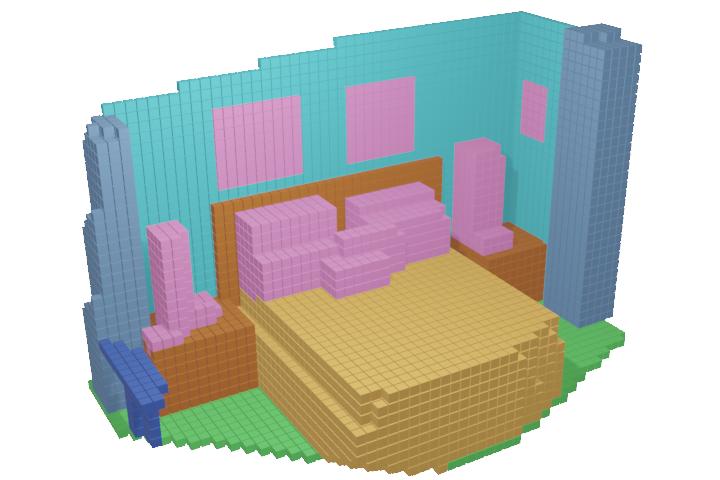}
\end{subfigure}
\begin{subfigure}{0.16\textwidth}
\includegraphics[trim=20 0 20 0, clip, width=\linewidth]{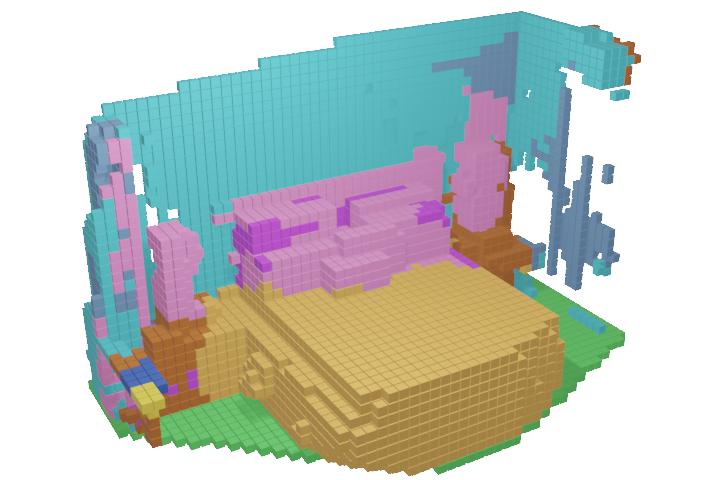}

\end{subfigure}
\begin{subfigure}{0.16\textwidth}
\includegraphics[trim=20 0 20 0, clip, width=\linewidth]{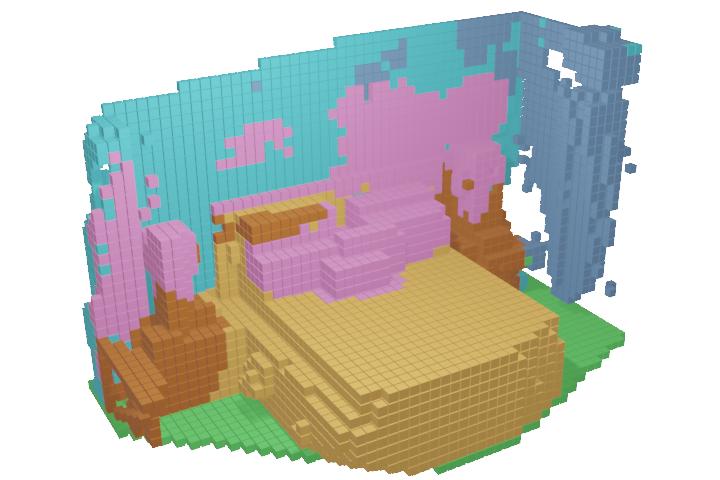}

\end{subfigure}
\begin{subfigure}{0.16\textwidth}
\includegraphics[trim=20 0 20 0, clip, width=\linewidth]{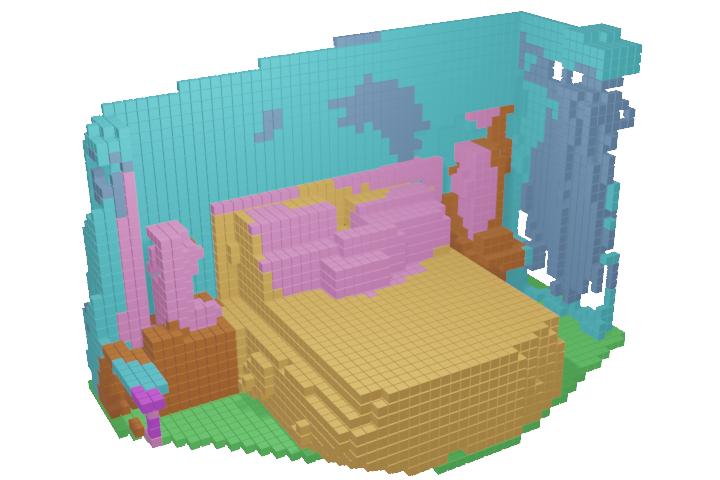}

\end{subfigure}
\begin{subfigure}{0.16\textwidth}
\includegraphics[trim=20 0 20 0, clip, width=\linewidth]{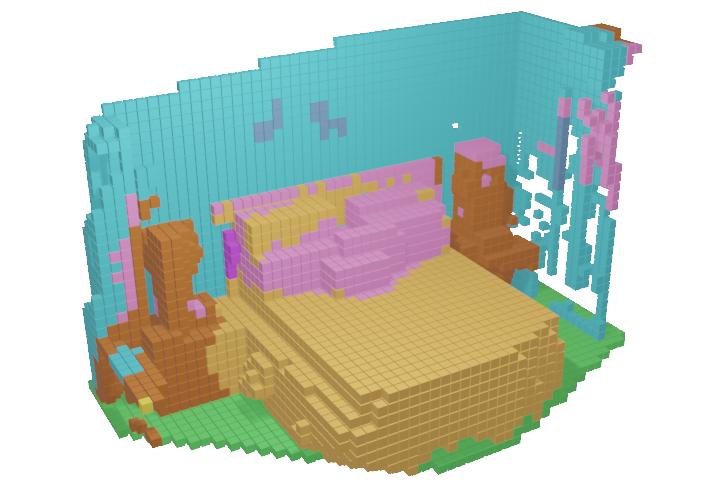}

\end{subfigure}


\begin{subfigure}{0.16\textwidth}
\includegraphics[width=\linewidth]{figures/NYU1202_0000_color.jpg}

\end{subfigure}
\begin{subfigure}{0.16\textwidth}
\includegraphics[trim=20 0 20 0, clip, width=\linewidth]{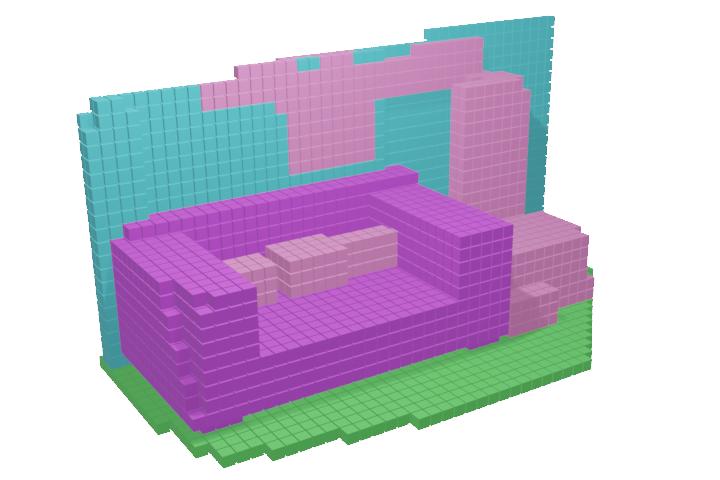}
\end{subfigure}
\begin{subfigure}{0.16\textwidth}
\includegraphics[trim=20 0 20 0, clip, width=\linewidth]{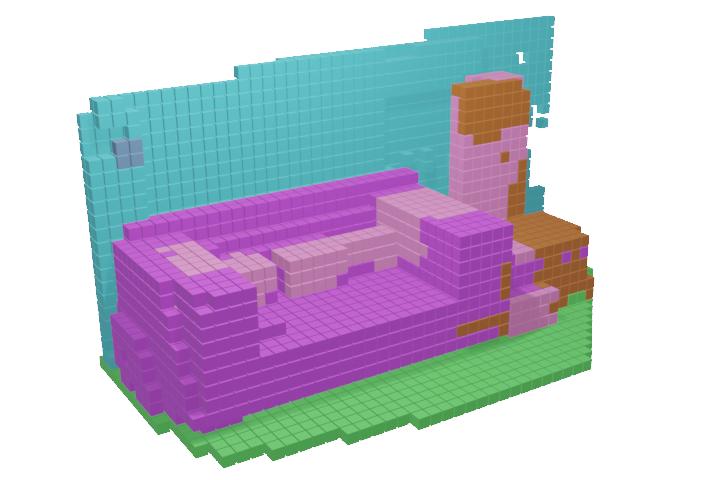}

\end{subfigure}
\begin{subfigure}{0.16\textwidth}
\includegraphics[trim=20 0 20 0, clip, width=\linewidth]{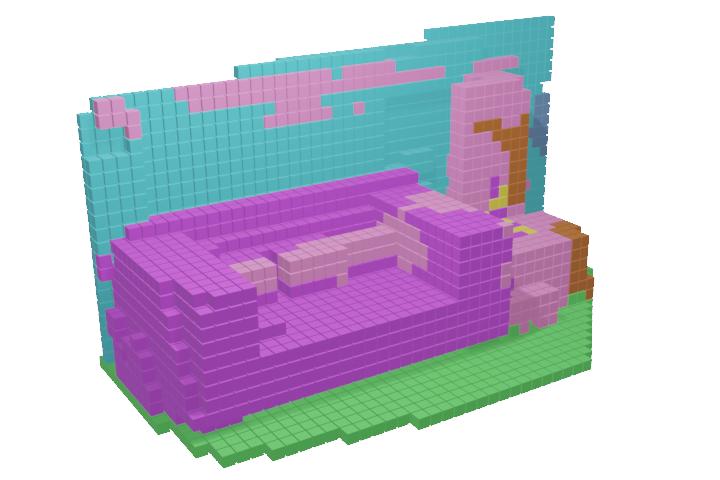}

\end{subfigure}
\begin{subfigure}{0.16\textwidth}
\includegraphics[trim=20 0 20 0, clip, width=\linewidth]{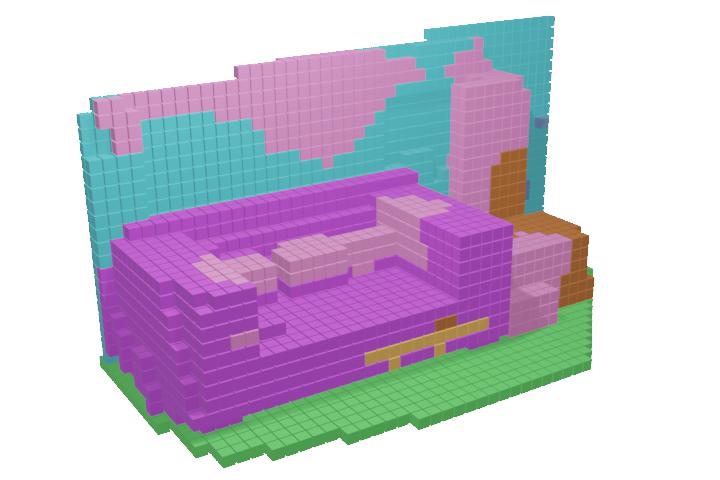}

\end{subfigure}
\begin{subfigure}{0.16\textwidth}
\includegraphics[trim=20 0 20 0, clip, width=\linewidth]{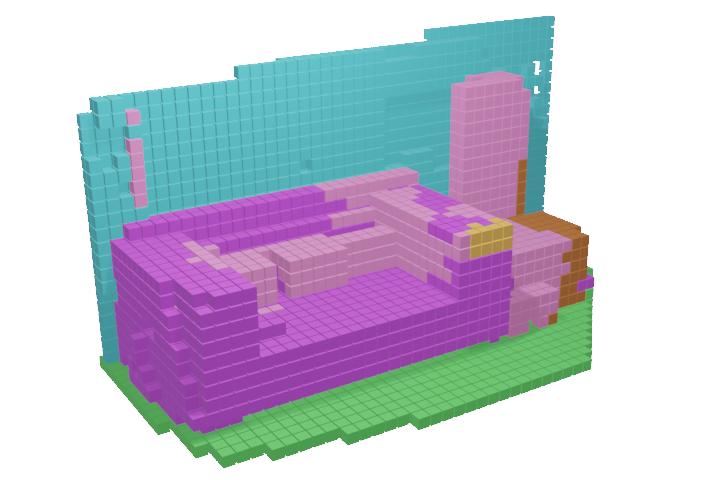}

\end{subfigure}


\begin{subfigure}{0.16\textwidth}
\includegraphics[width=\linewidth]{figures/NYU1314_0000_color.jpg}
\caption{RGB image}

\end{subfigure}
\begin{subfigure}{0.16\textwidth}
\includegraphics[trim=20 0 20 0, clip, width=\linewidth]{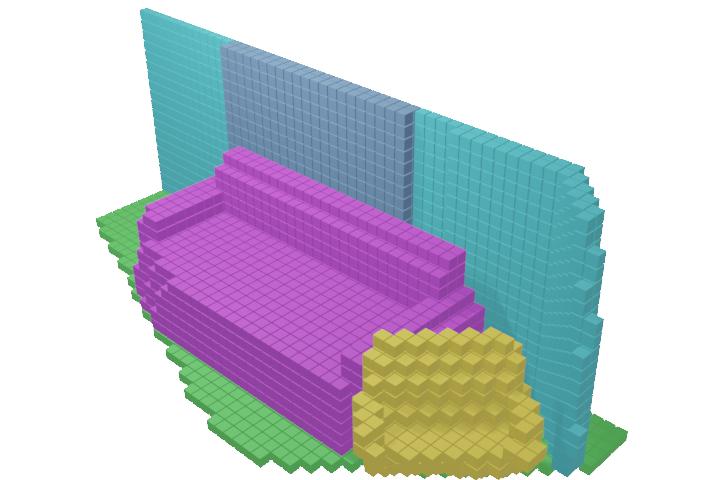}
\caption{Ground Truth}
\end{subfigure}
\begin{subfigure}{0.16\textwidth}
\includegraphics[trim=20 0 20 0, clip, width=\linewidth]{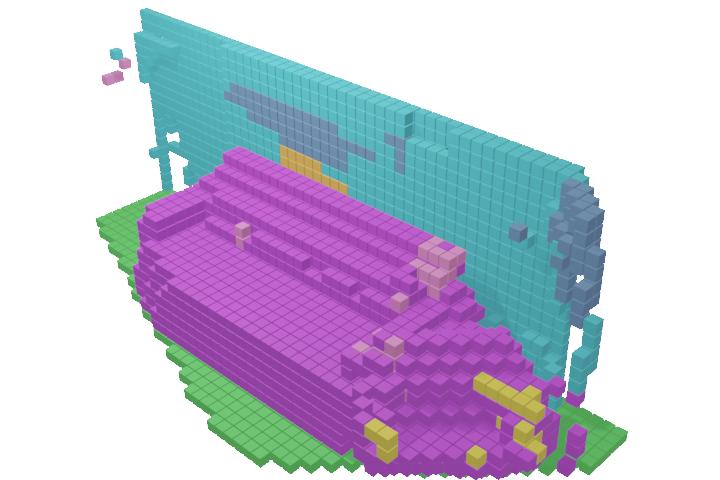}
\caption{SSCNet*}

\end{subfigure}
\begin{subfigure}{0.16\textwidth}
\includegraphics[trim=20 0 20 0, clip, width=\linewidth]{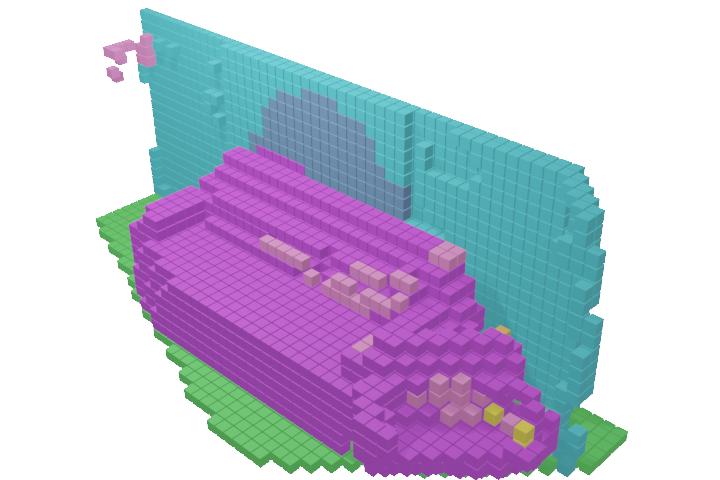}
\caption{EdgeNet-EF}

\end{subfigure}
\begin{subfigure}{0.16\textwidth}
\includegraphics[trim=20 0 20 0, clip, width=\linewidth]{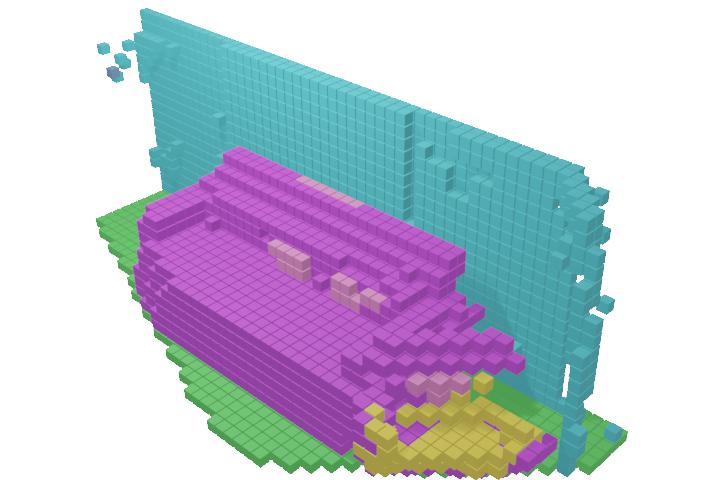}
\caption{EdgeNet-MF}

\end{subfigure}
\begin{subfigure}{0.16\textwidth}
\includegraphics[trim=20 0 20 0, clip, width=\linewidth]{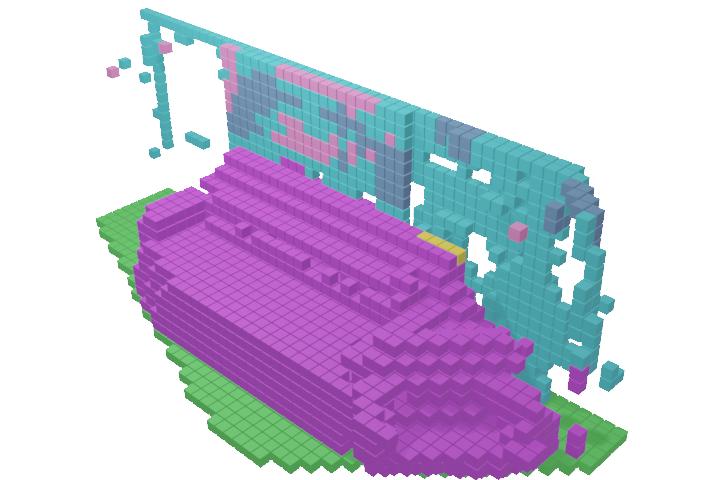}
\caption{EdgeNet-LF}

\end{subfigure}

\caption{\textbf{Qualitative Results}. We compare EdgeNet results using SSCNet* as a baseline on SUNCG and NYUDv2. Overall, EdgeNet gives more accurate voxel predictions, especially for hard to detect clases (best viewed in colour).}
\label{fig:qualy}
\end{figure*}

\subsection{Qualitative Results}
\label{sec:qualy}
Qualitative results on NYUDv2 are shown in Figure \ref{fig:qualy}.  Models used to generate the inferences were trained on SUNCG and fine tuned on NYUDv2.  We compare results of SSCNet* to our three models. It is visually perceptible that EdgeNet presents more accurate results.

In the first row of images of Figure \ref{fig:qualy}, note the presence of a picture and a window, and observe that the ground truth misses the window. SCCNet* did not detect the picture and the window while EdgeNet-MF detects the window and some parts of the picture. This ground truth mislabelling affects negatively the performance of EdgeNet.

The second row of Figure \ref{fig:qualy} also depicts some problems related to Ground Truth annotations on NYUDv2 dataset. Note that neither the papers fixed on the wall nor the shelf appear in the Ground Truth. All models captured the shelf, but only EdgetNet inferred the presence of objects fixed on the wall. When quantitative results are computed, these ground truth annotation flaws 
unfairly benefit the less precise models and harm more precise models like ours. 

\section {Discussion}
\label{sec:discussion}
 In this section we discuss key aspects and contributions of our proposed approach.

\textbf{Has the new training pipeline any influence over results?}
We compared the results originally achieved by SSCNet to results of the version of it trained with our pipeline (SSCNet*). On SUNCG we observed an  improvement of almost 20\% on semantic scene completion and more than 10\% on scene completion. Besides the improvements on model performance, the more computationally efficient pipeline also contributed to reduce training time from 7 days to 4 days when training on SUNCG and from 30 hours to 6 hours when training on NYUDv2, with a batch of size 3, whereas the original framework only allowed a batch size of 1 sample on a GTX 1080Ti (which has 11GB of memory). Besides reducing training time, larger batch sizes enhance training stability, acting as a regularizer \cite{smith_2018_batch_size}.  

\textbf{Is a deeper U-shaped CNN with dilated ResNet modules helpful?}
We investigated the effects of our architecture with and without aggregating edges. On both scenarios, our proposed architecture outperformed the shallower network, confirming that our network architecture is helpful.

\textbf{Is aggregating edges helpful? May Other 3D CNN architectures benefit from aggregating edges?}
We compared the original SSCNet architecture trained with our pipeline to a modified version of it that aggregates edges encoded with F-TSDF (SSCNet-E). SSCNet-E presented better results on SUNCG, demonstrating that the aggregation of edge information is helpful. We also observed improvements using a deeper depth-only network (EdgeNet-D). 
This experiments demonstrates that the proposed 3D volumetric representation of color edges can improve the performance of other previous depth only approaches.

\textbf{What is the best fusion strategy?}
The later the fusion, the higher is the memory requirement, due to the duplication of convolutional branches. Higher memory may imply in smaller batch sizes which may negatively impact learning. Liu \etal\cite{See_and_think_2018} observed better results using late fusion, but they faced the problem of higher memory consumption. Our choice was to fix the memory footprint, reducing the number of channels of duplicated branches without compromising the training time and stability.  However, very late fusion schemes may suffer from accuracy degradation due to reduced number of parameters in deeper layers. Taking those aspects into account, we found that a mid-level fusion strategy works and generalizes better for EdgeNet considering both synthetic and real datasets.

\textbf{How does EdgeNet compare to other RGB + depth approaches?} 
We have compared EdgeNet with other RGB + depth approachs on SUNCG (Table \ref{tab:2}) and NYUDv2 (Table~\ref{tab:1}. On SUNCG, EdgeNet versions surpassed previous approaches by a large margin. On NYU, EdgeNet got similar results as the solutions from  TNetFuse \cite{See_and_think_2018}, with less than 1\% difference. It is important to observe NYU ground truth annotations are not precise, which impacts negatively more accurate models. 
Another aspect that is worth mentioning is that TNetFuse needs a complex and less computationally efficient two-step training protocol, while EdgeNet and the previous depth-only solutions cited in this paper are end-to-end networks, with a much simpler and efficient training pipeline.

\section{Conclusion}
This paper presented a new approach to fuse depth and colour into a CNN for semantic scene completion. We introduced the use of F-TSDF encoded 3D projected edges extracted from RGB images. 
We also presented a new end-to-end network architecture capable of properly aggregating edges and depth, extracting useful information from both sources, without requiring previous 2D semantic segmentation training as is the case of previous approaches that combine depth and colour. Experiments with alternate models, showed that both aggregating edges and the new proposed architecture have positive impact on semantic scene completion, especially for hard to detect objects. Qualitative results show significant improvement for objects such as pictures, which cannot be differentiated by depth only. On SUNCG, we have achieved the best overall result, and on NYU, we  have achieved the state-of-the-art results of other approaches that use a more complex training protocol.

Experiments showed that our proposed approach of aggregating Edges may be applied to other existing solutions, opening room for further improvements. 

We also developed a lightweight training pipeline for the task, which reduced the memory footprint in comparison to other solutions and reduced the training time on SUNCG from 7 to 4 days and on NYUDv2 from 30 to 6 hours, that we intend to make public upon acceptance of this paper.






\bibliographystyle{IEEEtran}
\bibliography{paper}
%

\end{document}